\documentclass[runningheads]{llncs}
\usepackage{graphicx}
\usepackage{cite}
\usepackage{amsmath,amssymb,amsfonts}
\usepackage{algorithmic}
\usepackage{graphicx}
\usepackage{textcomp}
\usepackage{xcolor}
\usepackage{booktabs}
\usepackage{verbatim}
\usepackage{multirow}
\usepackage{makecell}
\usepackage{enumitem}

\newcommand{\Tau}{\mathrm{T}}
\begin{document}
\title{Bilinear Fusion of Commonsense Knowledge with Attention-Based NLI Models}
\titlerunning{BiCAM For Natural Language Inference}
%
\author{Amit Gajbhiye \and
Thomas Winterbottom \and
Noura Al Moubayed \and
Steven Bradley}
\authorrunning{Amit Gajbhiye et al.}
%
\institute{Department of Computer Science, Durham University, Durham, United Kingdom 
\email{\{amit.gajbhiye, thomas.i.winterbottom, noura.al-moubayed, s.p.bradley\}@durham.ac.uk}
}
\maketitle              
\begin{abstract}
We consider the task of incorporating real-world commonsense knowledge into deep Natural Language Inference (NLI) models. Existing external knowledge incorporation methods are limited to lexical-level knowledge and lack generalization across NLI models, datasets, and commonsense knowledge sources. To address these issues, we propose a novel NLI model-independent neural framework, BiCAM. BiCAM incorporates real-world commonsense knowledge into NLI models. Combined with convolutional feature detectors and bilinear feature fusion, BiCAM provides a conceptually simple mechanism that generalizes well.  Quantitative evaluations with two state-of-the-art NLI baselines on SNLI and SciTail datasets in conjunction with ConceptNet and Aristo Tuple KGs show that BiCAM considerably improves the accuracy the incorporated NLI baselines. For example, our BiECAM model, an instance of BiCAM, on the challenging SciTail dataset, improves the accuracy of incorporated baselines by 7.0\% with ConceptNet, and 8.0\% with Aristo Tuple KG.


\end{abstract}

\section{Introduction}
Natural Language Inference (NLI), also known as Recognizing Textual Entailment (RTE), is one of the key problems in the field of Natural Language Understanding (NLU). Popularised by a number of PASCAL RTE challenges, the task is formulated as a - ``directional relationship between pairs of text expressions, denoted by T (the entailing Text) and H (the entailed “Hypothesis”). Text T, entails hypothesis H, if humans reading T would typically infer that H is most likely true.'' \cite{rteBookDagan}. The task is very challenging as it requires an entailment system to acquire the linguistic knowledge (word meaning, syntactic structure and semantic interpretation), and also to understand commonsense knowledge.

In the context of artificial intelligence, commonsense knowledge is the set of background information about the everyday world, that an individual is expected to know or assume, and the ability to use it when appropriate \cite{minsky1986societyMind}. 
Many complex NLU applications such as machine reading \cite{leveraging2017Yang} achieved improved performance when supplied with commonsense knowledge.

\begingroup
\begin{table}[!h]
    \caption{SNLI example with commonsense triples (red) from ConceptNet KG.}
    \renewcommand{\arraystretch}{1.1}
    \centering
    \begin{tabular}{p{0.9\columnwidth}} \\ \hline

    \textbf{p:} Two young girls hang \textcolor{red}{tinsel} on a Christmas tree in a room with blue curtains. \textcolor{red}{(tinsel IsA decoration)} \\ \hline
     
     \textbf{h:} Two girls are decorating their Christmas \textcolor{red}{tree}. \textcolor{red}{(tree RelatedTo christmas)} \\ \hline
     
    \end{tabular}

    \label{tab:snliExamples}
\end{table}
\endgroup

Thus far, NLI research has not fully leveraged the additional information available via the use of commonsense knowledge. For example, state-of-the-art NLI models \cite{chen2018extknow, adventure2018khot} are limited to incorporating only lexical-level external knowledge, such as synonym and hypernymy. However, NLI is a complex reasoning task, in addition to lexical-level external knowledge, the task requires real-world commonsense knowledge to reason inference. Table \ref{tab:snliExamples} shows examples from the SNLI dataset \cite{snliData}, where the commonsense knowledge is retrieved from the ConceptNet Knowledge Graph (KG) \cite{conceptNet2017speer}. The common knowledge that, \textit{tinsel IsA decoration} and \textit{tree RelatedTo christmas} is useful to ascertain the inference relationship. 
Due to the lack of such common knowledge, state-of-the-art NLI models perform substantially worse for such premise-hypothesis pairs \cite{breakingNLI2018Glockner}. 

Incorporating external commonsense knowledge in deep neural NLI models is challenging. Existing models require considerable architectural changes with marginal performance gains \cite{chen2018extknow}. Incorporating such knowledge implicitly by refining word embeddings using KGs may negatively affect model performance \cite{dynamicintegration2017weissenborn}. Moreover, the existing external knowledge-based NLI models do not generalize well, and lack extensive evaluation across NLI datasets and KGs \cite{kang2018bridging}.

This paper aims to mitigate the aforementioned limitations. We present BiCAM (\textbf{Bi}linear fusion of \textbf{C}ommonsense knowledge with \textbf{A}ttention-based NLI \textbf{M}odels) - a novel neural network framework that incorporates NLI models without any architectural changes to the model. The BiCAM is NLI model-independent framework that generalizes across NLI models, datasets and commonsense knowledge sources. In the proposed framework, we first formulate the heuristics to retrieve commonsense knowledge from the KGs. We then embed retrieved knowledge with Holographic Embeddings (HolE) \cite{hole2016Nickel}, a KG embedding method to learn the embeddings of entities and relations in the KG. We learn the commonsense features from KG embeddings using a Convolutional Neural Network (CNN) based encoder. Finally, we use a state-of-the-art feature fusion technique, factorized bilinear pooling, to learn the joint representation of the learned commonsense features and the sentence features from the NLI model. 

Evaluation results on two established NLI baselines ESIM \cite{chen2017enhanced} and decomposable attention model \cite{parikh2016decomposable}, in combination with ConceptNet \cite{conceptNet2017speer} and Aristo Tuple \cite{aristoTuple2017Dalvi} KGs demonstrate that BiCAM considerably improve the accuracy of all the incorporated baselines. For example, compared with ESIM baseline, BiCAM achieves 7\% absolute improvement with ConceptNet and 8\% absolute improvement with AristoTuple KG on SciTail dataset. We analyze the effect of incorporating the different number of commonsense features and find that syntactically and semantically complex sentences require more commonsense knowledge to reason inference. Further, we evaluate the impact of various feature fusion techniques and demonstrate the efficacy of bilinear feature fusion. Finally, we analyze the examples from SNLI test set, where ESIM and BiCAM succeed and fail. 

In summary, the main contributions of this paper are: (1) We introduce the NLI model-independent neural framework, BiCAM, that generalizes across NLI models, datasets, and commonsense knowledge sources. (2) We devise an effective set of knowledge retrieval heuristics from KGs. (3) An extensive evaluation of the proposed approach with two established NLI baselines in combination with a general commonsense and (science) domain-specific KG on two NLI benchmarks.


\begingroup
\begin{figure}[!t]
  \centering
    \includegraphics[width=8cm,height=6.35cm]{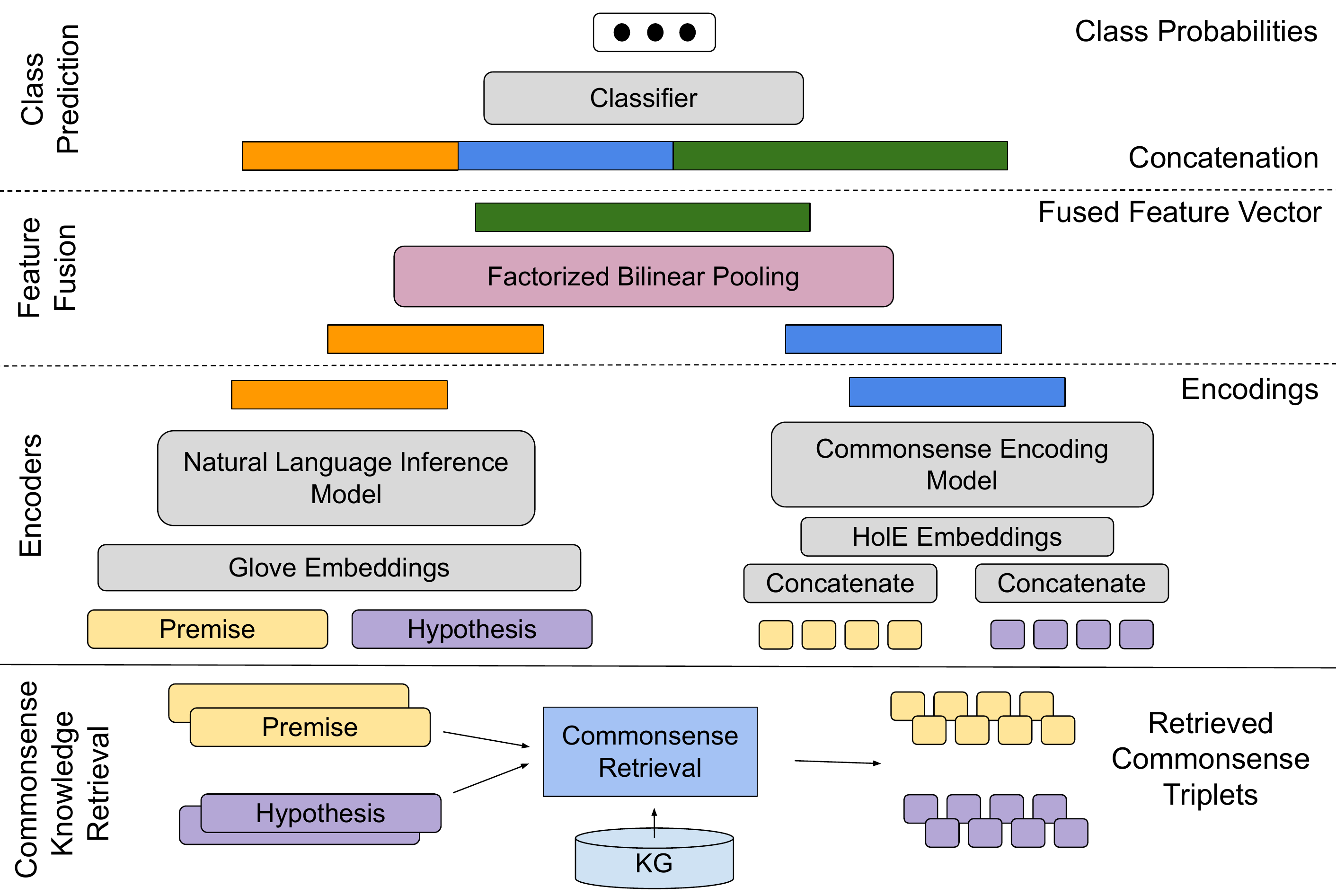}
    \caption{A high-level view of our proposed architecture (BiCAM). The data (premise, hypothesis and the corresponding commonsense triples) flows from bottom to top. Premise and the corresponding triples are depicted in yellow, hypothesis and the corresponding triples are shown in purple.}
    \label{fig:commonSenseModel}
\end{figure}
\endgroup

\section{Related Work}
Leveraging commonsense knowledge in NLU systems has long been proposed \cite{minsky1986societyMind}, however, NLI neural models have only recently started utilising commonsense knowledge. KIM \cite{chen2018extknow}, is the state-of-the-art neural Knowledge-based Inference Model, that incorporates lexical-level semantic knowledge into the attention and composition components. Specifically, external lexical knowledge (such as synonym and antonym) extracted from the lexical database, WordNet \cite{miller1995wordnet}, is used to form relation embeddings between premise-hypothesis words. The {A}dv{E}ntu{R}e \cite{adventure2018khot} framework train the decomposable attention model\cite{parikh2016decomposable} with adversarial training examples generated by incorporating knowledge from linguistic resources such as WordNet, and with a sequence-to-sequence neural generator. 
However, lexical knowledge individually is insufficient to reason about the premise-hypothesis relationship. Intuitively, when a human judges a premise-hypothesis relationship, a full range of real-world commonsense knowledge, and not just the lexical knowledge, is necessary to come to a conclusion \cite{minsky1986societyMind}. 
Therefore, we incorporate knowledge from and empirically evaluate BiCAM on the real-world commonsense KG, ConceptNet. We also evaluate BiCAM  on the (science) domain-specific KG, Aristo Tuple. 

NSnet \cite{kang2018bridging} is a neural-symbolic entailment model, that integrates the connectionist, deep learning approach with the symbolic approach for the scientific entailment task. The model decomposes each of the hypotheses into various facts and verifies each sub-fact against the premises using decomposable attention model and against the Aristo Tuple KB using a structured scorer. An aggregator network then combines the predictions from the two modules to get the final entailment score. Word embeddings are refined by dynamically incorporating relevant background knowledge from external knowledge sources in \cite{dynamicintegration2017weissenborn}. Our approach differs in the manner and the level at which commonsense is incorporated. We fuse the commonsense features to the sentence encodings of the premise and hypothesis which we show achieves a better performance.

\section{Methods}
A high-level view of our proposed BiCAM framework is illustrated in Figure \ref{fig:commonSenseModel}. In this section, we discuss the individual BiCAM components and the uniquely structured framework. 

\subsection{Commonsense Knowledge Retrieval}
\label{sec:comSenseExt}
To extract external commonsense knowledge we consider two KGs: ConceptNet, for general real-world commonsense knowledge and Aristo Tuple, for (science) domain-specific knowledge. The knowledge in these KGs is represented as a triple $(head,\;relation,\;tail)$, where $head$ and $tail$ are the real-world entities and the $relation$, is a specific set of associations, describing the relation between entities. For example, $(tinsel\;IsA\;decoration)$ is a triple in ConceptNet KG.

Retrieval and preparation of contextually specific and relevant information from knowledge graphs are complex and challenging tasks and is the crucial step in our model. We use a heuristic retrieval mechanism for knowledge retrieval. We find empirically that non-specific commonsense knowledge from the KGs degrades the model performance. Heuristic mechanism is fast and is effective in filtering irrelevant knowledge. We formulate the following heuristics and illustrate the triples retrieved by the application of each heuristic in Table \ref{tab:extractionRules}.

\begingroup
\begin{table*}[ht]
\caption{A step by step illustration of commonsense knowledge retrieval for a SNLI premise-hypothesis pair from ConceptNet. Step 4 shows the final set of triplets for the premise and hypothesis.}
    \renewcommand{\arraystretch}{1.20}
    \centering
    \resizebox{\columnwidth}!{ 
    \begin{tabular}{c|p{7cm}|l} \hline

         \textbf{Step} & \textbf{Premise} & \textbf{Hypothesis} \\ \hline
         
        Input & A \textbf{white horse} is pulling a \textbf{cart} while a \textbf{man} stands and watches. & An \textbf{animal}                   is walking \textbf{outside}. \\ \hline
        1. & (`white', `horse', `pulling', `cart', `man', `stands', `watches') & (`animal', `walking', `outside'')  \\ \hline
        
        2. & (horse  has\_property white),\;(cart related\_to horse) &  (animal at\_location outside) \\ \hline
        
        3. & \makecell[l]{(horse is\_a animal),\;(horse related\_to animal), \\ (horse at\_location outside)} 
           & \makecell[l]{(animal related\_to horse),\;(animal antonym man), \\ (animal distinct\_from man)} \\\hline
                                
        4. & \makecell[l]{(horse  has\_property white),\;(cart related\_to \\ horse)
        (horse is\_a animal),\;(horse at\_location \\ outside)} & \makecell[l]{(animal at\_location outside),\;(animal related\_to horse) \\ (animal antonym man)} \\ \hline

    \end{tabular} 
}
    \label{tab:extractionRules}
\end{table*}
\endgroup

\begin{enumerate}[noitemsep,nolistsep]
    \item Stop words are removed from the premise and hypothesis.
    
    \item To identify the relations between the words within the premise or hypothesis, we retrieve all triples involving each pair of words as head and tail. 
    
    \item To identify the relations from premise words to hypothesis words, we retrieve the triples with premise words as head and the words of hypothesis as tail. For hypothesis, we extract the relations from the hypothesis to premise.
    
    
    \item The relation \textit{RelatedTo} has the largest number of triples in ConceptNet. Although the relation communicates that the head and tail are related, it does not specify the specific relationship between them. To eschew the extracted commonsense knowledge from non-specific information and a higher number of triples with \textit{RelatedTo} relation, we randomly select one triplet with \textit{RelatedTo} relation, if multiple such triples are extracted. Additionally, we removed any duplicated triples from the final set of retrieved triples.
    
    \item Finally, if the words of the premise and the hypothesis do not extract any commonsense knowledge by the application of above heuristics, we randomly select a word from them and extract a triple from one of the relations in $(entails,\;synonym,\;antonym)$.
\end{enumerate}

\subsection{Encoders}
\label{subSec:encoders}
\textbf{Commonsense Encoding Model.} The model learns the features from the retrieved commonsense triples. We provide a layer-by-layer description.

\textbf{Embedding Layer.} We learn the Holographic Embeddings (HolE) \cite{hole2016Nickel} of KG triples. Given a commonsense triple $(h,\;r,\;t)$, HolE represents both the entities and relations as vectors in $\mathbb{R}^d$. First, HolE compose the head and tail into $\mathbf{h} \star \mathbf{t} \in \mathbb{R}^d$ using the circular correlation:
\begin{equation}
\label{eq:circularCorrelation}
[\mathbf{h}\star \mathbf{t}]_i = \sum_{k=0}^{d-1} [\mathbf{h}]_k \odot [\mathbf{t}_{(k+i) \mathrm{mod}\;d}]
\end{equation}
where $\odot$ denotes the Hadamard product. The compositional vector obtained is then matched with the continuous representation of relation to score the commonsense triple using the scoring function defined as: 
\begin{equation}
\label{eq:relationScoring}
f_r(h, t) = \mathbf{r}^\mathrm{T}(\mathbf{h}\star \mathbf{t}) = \sum_{i=0}^{d-1} [\mathbf{r}]_i \sum_{k=0}^{d-1} [\mathbf{h}]_k \odot [\mathbf{t}]_{(k+i)\mathrm{mod}\;i}
\end{equation}
where $\mathbf{r} \in \mathbb{R}^d$ is the relation embedding. The score measures the plausibility of the commonsense triple. We train the HolE embeddings ($\Theta$) using the pairwise ranking loss computed as: 
\begin{equation}
\label{eq:lossFunction}
    \min_{\Theta} \sum_{i \in \Gamma_{+}} \sum_{j \in \Gamma_{-}} \max(0\\, \; \gamma + \sigma(\eta_{j}) - \sigma(\eta_{i}))
\end{equation}
where $\Gamma_{+}$ denotes the set of triples in the KG, $\Gamma_{-}$ denotes the ``negative" triples that are not observed in KG and $\gamma > 0$ specifies the width of margin, $\sigma(.)$ denotes the logistic function and $\eta$ is the value of the scoring function.  

For ConceptNet and Aristo Tuple, we train the HolE embeddings for the triples retrieved from the SNLI and SciTail vocabulary. We use AdaGrad \cite{adagrad2011duchi} to optimize the objective in Eq \ref{eq:lossFunction}, via an extensive grid search over an initial learning rate of $(0.001,0.01,0.1)$, a margin of $(0.2, 1, 2, 10)$, mini-batch size $(50, 100, 150, 200)$ and entity embedding dimensions of $(50, 100, 150, 200)$. At each gradient step, we randomly generate $5$ negative $tail$ entities with respect to a positive triple. The learned HolE embeddings are evaluated on the triplet classification task. For SNLI/ConceptNet pair, the model achieves the highest accuracy of $64.0\%$ with an embedding dimension of $150$. For SciTail/ConceptNet and SciTail/Aristo Tuple pairs, HolE reported the top accuracy of $62.8\%$ and $69.4\%$ respectively at embedding dimension $100$. 

\textbf{Encoding Layer.} To learn the features over the pre-trained HolE embeddings, we employ a CNN-based neural model \cite{ccnForSentClassi2014Kim}. 

For each premise/hypothesis, let $\Tau = (\tau_1,\tau_2, \ldots, \tau_m)$ be a sequence of length $n$ created by joining the $m$ retrieved triples from the KG. Each $\tau$ is of the form $(h, r, t)$ and, hence, $n=3m$. The sequence $\Tau$, padded where necessary, and represented as:
\begin{equation}
\label{eq:tripletSeq}
    \Tau = ({x}_1, {x}_2, , {x}_3), ({x}_4, {x}_5, , {x}_6),  \ldots, ({x}_{n-2}, {x}_{n-1}, {x}_n)
\end{equation}
where, $x_{i}$ is the \textit{i}-th word in the sequence. Let $\mathbf{{x}}_i \in \mathbb{R}^d$ be the $d$-dimensional pre-trained HolE embedding corresponding to the $i$-th word. A sentence of length $n$ is represented as a matrix $\mathbf{X} \in \mathbb{R}^{d \times n}$, by concatenating its word embeddings as columns, \textit{i.e.}, $\mathbf{{x}}_i$ is the $i-$th column of $\mathbf{X}$. We apply a convolution operation with filter $\mathbf{W} \in \mathbb{R}^{d \times h}$, to a window of $h$ words. The convolution operation learns a new feature map from the set of $h$ words with the operation:
\begin{equation}
\label{eq:featureGeneration}
    c = f(\mathbf{X} * \mathbf{W} + b) \in \mathbb{R}^{(\frac{n-h}{s})+1}
\end{equation}
where, $b \in \mathbb{R}^{(\frac{n-h}{s})+1} $ is the bias term, $s$ is the stride of convolution filter, and $f(\cdot)$ is the activation function, rectified linear unit in our experiments and $*$ denote convolution operation. The filter convolve over each window $\mathbf{(x}_{ih +1 \colon (i+1)h})$ where $0 \leq i \leq n-1$ in $\mathbf{X}$. We set the $h$ and $s$ to $3$ for the commonsense triples. Convolving the same filter with the $3$-gram beginning at every $3^\mathrm{rd}$ position in the triple sequence allows the features to be extracted from every triplet from the KG. We then apply a max-over-time pooling operation over the feature map and take the maximum value $\hat{c} = \mathrm{max} \{\mathbf{c}\}$ as a feature corresponding to this filter. Max pooling operation captures the most important feature for each feature map. 

Above we detailed the process of extracting one feature from one filter. Multiple filters (with fixed window size and stride of 3) are employed to obtain multiple features.
Each filter is considered as a linguistic feature detector that learns to recognize a specific feature from the commonsense triple. The output of the commonsense encoder is a $l$-dimensional vector to represent commonsense. 

\textbf{NLI Encoders.} 
We incorporate BiCAM with two established NLI baselines: ESIM \cite{chen2017enhanced} and decomposable attention model \cite{parikh2016decomposable}. 

\textbf{Feature Fusion.}
\label{subsubSec:bilinerPool}
We apply factorized bilinear pooling \cite{bilinearPool2017Zhou} to fuse the commonsense features and NLI sentence features. Let $\mathbf{p}$ and
$\mathbf{h}$ be the NLI model generated encoding of premise and hypothesis. Also, let $\mathbf{p}_{cs}$ and $\mathbf{h}_{cs}$ denote the corresponding commonsense encoding generated by commonsense encoding model. We apply the factorized bilinear pooling defined as: 
\begin{equation}
\label{eq:bilinearPool}
\begin{aligned}
    \mathbf{z}_{p} &= \mathrm{SumPooling}(\widetilde{U} \mathbf{p} \odot \widetilde{V} \mathbf{p}_{cs}, k) \\
    \mathbf{z}_{h} &= \mathrm{SumPooling}(\widetilde{U} \mathbf{h} \odot \widetilde{V} \mathbf{h}_{cs}, k),
\end{aligned}
\end{equation}
where SumPooling$(x, k)$ denote a sum pooling over $x$ with a one dimensional non-overlapped window of size $k$, $\widetilde{U}$ and $\widetilde{V}$ are projection matrices learned during training, $\odot$ is the Hadamard product and $z$ is the fused feature vector. To prevent overfitting, we also added a dropout layer \cite{dropoutnli2018gajbhiye} after the element-wise multiplication of the projection matrices. Further, to allow the model to converge to a satisfactory local minimum, we append power normalization ($\mathbf{z} \leftarrow \mathrm{sign} (\mathbf{z})|\mathbf{z}|^{0.5}$) and $l_{2}$ normalization layers ($\mathbf{z} \leftarrow \mathbf{z}/\|\mathbf{z}\|$) after SumPooling layer \cite{bilinearPool2017Zhou}.
The factorized bilinear pooling captures the complex association between the features from premise-hypothesis and the corresponding commonsense features. The pooling method is implemented as a feed-forward neural network.

\textbf{Classification Layer.} We classify the relationship between premise and hypothesis using a Multilayer Perceptron (MLP) classifier. The input to the MLP is the concatenation of sentence encodings ($\mathbf{p}$ and $\mathbf{h}$) obtained from NLI model and the corresponding encodings ($\mathbf{z_p}$ and $\mathbf{z_h}$) obtained from feature fusion layer.
The MLP consists of two hidden layers with $tanh$ activation and a softmax output layer to obtain the probability distribution for each class. The network is trained in an end-to-end manner using multi-class cross-entropy loss.

\section{Experiments and Results}
Our aim is to incorporate commonsense knowledge into NLI models in order to augment the reasoning capabilities. The method should generalize across different NLI datasets, models and KGs. We evaluate BiCAM using two attention-based NLI baselines on two benchmarks in combination with two KGs. We compare our models with both external knowledge-based and attention-based NLI models. We refer to BiCAM as Bi\textbf{D}CAM, when the decomposable attention model is used as NLI baseline and Bi\textbf{E}CAM, when ESIM is used (see Figure \ref{fig:commonSenseModel}).

\textbf{Datasets.} We assess  \textbf{BiCAMs} (BiDCAM and BiECAM) on \textbf{SNLI} (570K examples) and \textbf{SciTail} (27K examples) benchmarks. We consider \textbf{ConceptNet} for general commonsense, and \textbf{Aristo Tuple} for domain-specific knowledge. 

\textbf{Results on SNLI.}
Table \ref{tab:snliResult} shows the results of the state-of-the-art external knowledge-based and attention-based NLI models in comparison to BiCAMs. We evaluate ConceptNet KG for commonsense knowledge for the SNLI dataset. The models, BiDCAM and BiECAM, improve the performance of their respective attention-based baselines (decomposable attention and ESIM models) by +0.4\% and +0.8\%. 
BiCAMs also perform consistently better among the external knowledge-based and attention-based NLI models. BiECAM model achieves an accuracy of 88.8\% competitive to the state-of-art external knowledge-based NLI models, ESIM+Syntactic Tree LSTM \cite{chen2017enhanced} and KIM \cite{chen2018extknow} without any architectural changes to the underlying NLI models.
\begin{table}
    \caption{NLI Models: Test accuracy. For our models, BiCAMs,  the percentage in the parenthesis shows the performance improvement over the base models.}
\parbox{.50\linewidth}{
\centering

    \resizebox{0.5\columnwidth}!{
    \begin{tabular}{lc} \hline
        
        \multicolumn{2}{c}{\textbf{SNLI Dataset}} \\ \hline
        \textbf{NLI Model} & \textbf{Test Acc(\%)} \\ \hline 
        
         \multicolumn{2}{l}{\textbf{External Knowledge-based Baselines}} \\ \hline
         {A}dv{E}ntu{R}e \cite{adventure2018khot} & 84.6 \\ 
         BiLSTM ($\mathrm{E}_3$) \cite{dynamicintegration2017weissenborn} & 86.5 \\ 
         ESIM ($\mathrm{E}_3$) \cite{dynamicintegration2017weissenborn} & 87.3 \\
         Char+CoVe-L \cite{cove2017mccann} & 88.1 \\
         ESIM + Syntactic TreeLSTM \cite{chen2017enhanced} & \textbf{88.6} \\
         KIM \cite{chen2018extknow} & \textbf{88.6} \\ \hline

         \multicolumn{2}{l}{\textbf{Attention-based Baselines}} \\ \hline
         CAM \cite{gajbhiye2018CAM} & 86.1 \\
         Decomposable Attention \cite{parikh2016decomposable}& \textbf{86.3} \\
         ESIM \cite{chen2017enhanced} &   \textbf{88.0} \\ \hline

         \multicolumn{2}{l}{\textbf{Our Models}} \\ \hline
          BiDCAM + ConceptNet  & \textbf{86.7}  (\textbf{+0.4\%}) \\
         BiECAM + ConceptNet & \textbf{88.8}  (\textbf{+0.8\%}) \\ \hline
    \end{tabular}
    \label{tab:snliResult}
    }
}%
\hfill
\parbox{.50\linewidth}{
\centering
    \resizebox{0.50\columnwidth}!{
    \begin{tabular}{lc} \hline

        \multicolumn{2}{c}{\textbf{SciTail Dataset}} \\ \hline
        \textbf{NLI Model} & \textbf{Test Acc\%)} \\ \hline

         \multicolumn{2}{l}{\textbf{External Knowledge-based Baseline}} \\ \hline
         Majority classifier \cite{kang2018bridging} & 60.3 \\
         {A}dv{E}ntu{R}e(seq2seq)\cite{adventure2018khot} & 76.9 \\ \hline
         
         
         \multicolumn{2}{l}{\textbf{Attention-based Baseline}} \\ \hline
         ESIM \cite{chen2017enhanced}&   \textbf{70.6} \\
         Decomposable Attention \cite{parikh2016decomposable} & \textbf{72.3} \\ 
         CAM \cite{gajbhiye2018CAM} & 77.0 \\
         DGEM \cite{scitail} & 77.3 \\ \hline

         \multicolumn{2}{l}{\textbf{Our Models}} \\ \hline
         BiDCAM + ConceptNet & \textbf{76.8}  (\textbf{+4.5\%})   \\
         BiDCAM + Aristo Tuple & \textbf{77.3}  (\textbf{+5.0\%}) \\ 
         
         BiECAM + ConceptNet & \textbf{77.6}  (\textbf{+7.0\%}) \\
         BiECAM + Aristo Tuple & \textbf{78.6}  (\textbf{+8.0\%}) \\ \hline
        
    \end{tabular}
    }
    \label{tab:sciTailResults}
}
\end{table}

\textbf{Results on SciTail.} The test accuracy of different NLI models on SciTail benchmark is summarised in Table \ref{tab:sciTailResults}. For SciTail, we study the performance of BiCAMs on the general commonsense ConceptNet KG as well as the (science) domain-targeted Aristo Tuple KG. All our models significantly outperform the incorporated baselines across both the KGs, achieving absolute improvements of up to 4.5\% (BiDCAM + ConceptNet), 5\% (BiDCAM + Aristo Tuple) on decomposable attention baseline and 7\% (BiECAM + ConceptNet), 8\% (BiECAM + Aristo Tuple) on ESIM baseline. This demonstrates our framework's ability to generalize well across a number of NLI models and different KGs. 
All our models perform competitively on attention-based baselines, CAM and DGEM. BiECAM + Aristo Tuple observes an accuracy improvement of 1.3\% over the previous state-of-the-art DGEM model.

\section{Analysis}
\subsection{Number of Commonsense Features}
\label{sec:NumCommonsenseFeature}
\begin{figure}[!h]
    \centering
    \includegraphics[width=0.80\textwidth]{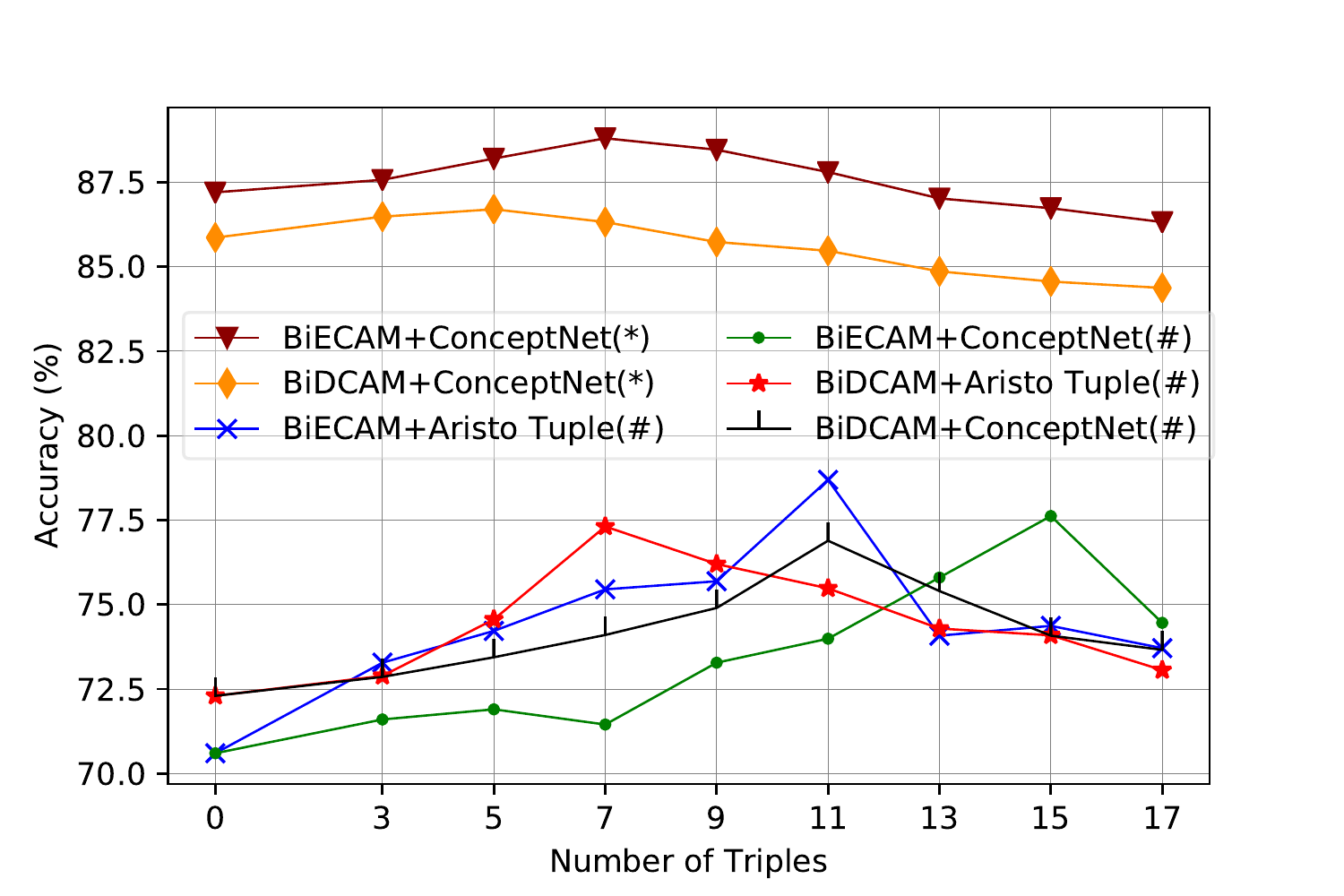}
    \caption{Accuracy of BiCAMs with varying amount of commonsense triples. (*) denotes SNLI and (\#) SciTail datasets.}
    \label{fig:lenPlots}
\end{figure}

To investigate the effect of incorporating various numbers of commonsense features, we vary the number of triples input to the commonsense encoding model. Particularly, we are interested in answering the question: How many commonsense features are required for optimal model performance? Figure \ref{fig:lenPlots} shows the results of the experiment.

\textbf{For SNLI}, the model BiECAM + ConceptNet achieves the highest accuracy (88.8\%) using 7 triples. We observe a decrease in accuracy with increasing the number of triples. BiDCAM + ConceptNet follow the same trend, however, it attains the highest accuracy (86.7\%) with the fewer number (5) of triples. The fewer number of triples required for BiCAMs to achieve their maximum accuracies on SNLI dataset, is attributed to the limited linguistic variation and short average length of stop-word filtered premise (7.35 for entails and neutral class) and hypothesis (3.61 for entails and 4.45 for neutral class) \cite{scitail} of the SNLI dataset, which limit its ability to fully extract and exploit KG knowledge.

\textbf{For SciTail}, the BiCAMs, when evaluated using the general commonsense knowledge source ConceptNet, require a relatively high number of triplets (11 and 15 resp.) to achieve their maximum accuracy. This is due to the higher syntactic and semantic complexity of SciTail, that needs more knowledge to reason inference. However, when evaluated with the domain-specific Aristo Tuple KG, the models achieve the highest accuracies with fewer (BiDCAM at 7 and BiECAM at 11) triples. The specialised scientific knowledge in Aristo Tuple improves the model performance with less external knowledge.

We observe that the BiCAMs, when trained on SciTail dataset, require a higher number of triples to attain maximum accuracy relative to when trained on the SNLI dataset. This can be attributed to the small training size of the SciTail dataset, which thus requires a higher number of triples to compensate for missing knowledge. We conclude that: (1) The commonsense features, when incorporated in the correct number, help reason the relationship between premise and hypothesis. (2) The number of commonsense features required depends on the syntax, semantics and size of the target dataset, as well as the domain of source KG. 

\subsection{Ablation Study}
To evaluate the impact of factorized bilinear feature fusion, we perform an ablation study on BiECAM + Aristo Tuple, our best performing model on the SciTail dataset. Table \ref{tab:ablationResults} demonstrates the performance of various non-bilinear and bilinear pooling methods. We observe that factorized bilinear pooling significantly outperforms all the non-bilinear pooling methods. To ascertain that the performance gain is not due to the higher number of parameters in bilinear method, we stack fully connected layers (with 1200 units per layer, ReLU activation and dropout) to increase the parameters in non-bilinear methods. We observe that increasing the number of parameters does not increase the model accuracy. The high accuracy of factorized bilinear pooling may be attributed to the outer product between the NLI sentence and the commonsense feature vectors. Outer product allows each feature point in the two feature vectors to interact and capture associations between them. The joint representations created in such manner are more expressive than the representations created through concatenation or element-wise summation or multiplication.

\begin{table}
\parbox{.45\columnwidth}{
\centering
    \caption{Ablation Study. ($\odot$ implies Elementwise)}
    \label{tab:ablationResults}
    
    \begin{tabular}{lc} \hline

       \textbf{Fusion Method}  & \textbf{Acc(\%)} \\ \hline \hline
       Concat &  74.6 \\
       FC + Concat & 75.5 \\
       FC + FC + Concat & 74.3 \\ \hline

       FC + $\odot$ Sum  & 72.5 \\
       FC + FC + $\odot$ Sum & 73.3 \\ \hline

       FC + $\odot$ Product & 76.4 \\
       FC + FC + $\odot$ Product & 76.8 \\ \hline
       
       \makecell[l]{FC + $\odot$ Difference Concat\\ FC + $\odot$ Product}  & 77.6 \\ \hline
       
       Factorized Biliniear Pooling & \textbf{78.6} \\ \hline
    \end{tabular}
    }
\hfill
\parbox{.50\columnwidth}{
\caption{Qualitative Analysis}
\label{tab:qualitativeAnalysis}
\centering
    \renewcommand{\arraystretch}{1.2}
    \begin{tabular}{p{0.5\columnwidth}} \hline
        \textbf{BiECAM Correct ESIM Incorrect} \\ \hline 
        
        \textbf{p:} Four boys are about to be \textbf{hit} by an  approaching \textbf{wave}.
        \textit{(wave RelatedTo crash)} \\
        \textbf{h:} A giant \textbf{wave} is about to \textbf{crash} on some boys.
        \textit{(crash IsA hit)} \\ \hline
        
        \textbf{BiECAM Incorrect ESIM Correct} \\ \hline 
        
        \textbf{p:} A \textbf{red} truck is parked next to a burning \textbf{blue} building while a man in a \textbf{green} vest runs toward it.
        \textit{(red Antonym blue), (blue Antonym green), (green Antonym red)} \\ 
        \textbf{h:} The burning \textbf{blue} building smells of smoke.
        \textit{(blue Antonym red), (blue Antonym green)} \\ \hline
    \end{tabular}
    }%
\end{table}

For the commonsense encoder, our experiments with Recurrent Neural Networks (RNNs), LSTMs and BiLSTMs, considerably degraded the performance of the BiCAMs. This may be attributed to the inherent nature of RNNs, which learns the representations of words in the context of all previous words in the sequence. However, the set of triples input to the commonsense encoder is sequential within an individual triple. For example, in the set of triples - \textit{(outside Antonym inside) and (table RelatedTo eating)}, the word \textit{inside} is associated with the words in its own triple, \textit{outside} and \textit{Antonym}, but not with the words \textit{table}, \textit{RelatedTo} and \textit{eating} of the second triple. RNNs, due to their inherent recurrent nature, learn the incorrect features from the part-sequential input of set of triples. In contrast, CNNs learns features independently of the position of words in the sequence. In the commonsense encoder, learning the features over the window of three words with a stride of three, allows the correct features to be learnt from the part-sequential set of input triples.

\subsection{Qualitative Analysis}
Table \ref{tab:qualitativeAnalysis} highlights selected sentences from the SNLI test set showing correct and incorrect inference prediction example for both BiECAM and the baseline ESIM. For the first example, BiECAM has additional context for premise and hypothesis from the knowledge that \textit{(wave RelatedTo crash)} and \textit{(crash IsA hit)}, which helps the model to correctly predict the inference class. However, the specific knowledge, about the \textit{wave} and the \textit{crash} is not available to the baseline ESIM model and hence, it incorrectly predicts the inference class. 

We observe that BiECAM fails to predict the correct inference class when noisy and irrelevant knowledge is retrieved from the KGs. For example, the last test case in Table \ref{tab:qualitativeAnalysis}, only retrieves the information that colors (such as red and blue) are antonyms of each other. The retrieved knowledge is irrelevant and is not completely correct, which does not help BiECAM.

\section{Conclusions}
We have introduced an NLI model-independent neural framework, BiCAM, that incorporates commonsense knowledge to augment the reasoning capabilities of NLI models. Combined with convolutional feature detectors and bilinear feature fusion, BiCAM provides a conceptually simple mechanism that generalizes across NLI models, datasets and KGs. Moreover, BiCAM can be easily applied to different NLI model and KG combinations. Evaluation results show that our BiCAM considerably improves the performance of all the NLI baselines it incorporates, and does so without any architectural change to the incorporated NLI model. BiCAM achieves state-of-the-art performance on SNLI with ConceptNet KG, outperforming existing state-of-the-art external knowledge-based NLI models. Particularly for the smaller, syntactically and semantically complex SciTail dataset, commonsense knowledge incorporation via BiCAM achieves performance improvements of 7.0\% with ConceptNet and 8.0\% with Aristo Tuple KG. Further analysis shows that the sufficient number of commonsense features required depends upon the syntax, semantics and size of the target dataset, as well as the domain of source KG. We observe that retrieval and selection of commonsense knowledge relevant for inference is challenging. In future work, we plan to leverage contextual word embeddings for commonsense knowledge retrieval from KGs.

%
%
%
\bibliographystyle{splncs04}
\bibliography{emnlp2018}
\end{document}